# A Probabilistic Reasoning Environment

Kathryn Blackmond Laskey
Decision Science Consortium, Inc.
1895 Preston White Drive, Reston, VA 22091

**Abstract**

A framework is presented for a computational theory of probabilistic argument. The Probabilistic Reasoning Environment encodes knowledge at three levels. At the deepest level are a set of schemata encoding the system's domain knowledge. This knowledge is used to build a set of second-level arguments, which are structured for efficient recapture of the knowledge used to construct them. Finally, at the top level is a Bayesian network constructed from the arguments. The system is designed to facilitate not just propagation of beliefs and assimilation of evidence, but also the dynamic process of constructing a belief network, evaluating its adequacy, and revising it when necessary.

## 1 Introduction

Attempts over the past thirty years at automating intelligence have altered our view of what it means to be intelligent. Computers can now beat all but the very best human chess players, but we remain very far from automating the kind of everyday reasoning that requires no special intellectual ability. We are discovering just how impressive ordinary intelligence is.

Ordinary intelligence means the ability to attack an ill-formulated problem, building a problem formulation as we try to solve the problem. It means being able to act in a world that is only partially understood. It means allocating our effort between solving the problem using the model we have and improving our model. It means dealing with uncertainty not just about which outcomes will occur within a given problem structure but also about what the problem structure is.

Dealing with uncertainty is at the heart of intelligent reasoning. We need to be able to plan and take actions that depend on outcomes we can predict only imperfectly. We make these predictions using a model of the world that we know is inadequate. We use inaccurate models for computational reasons, or because they are all we have. Frequently the world surprises us by failing to conform to our expectations. When this happens, we try to diagnose the cause of our failure to predict. Was the failure drastic enough to require changing our model, or can the problem be swept under the rug (until too many "insignificant" failures accumulate)? How shall we change the model? Is minor tweaking sufficient, or is wholesale revision needed? Can we think of a plausible alternative model, or do we remain in confusion?

Our current world model drives this revision process. We are more likely to ignore failures when there is no alternate model to explain them. Sometimes we cling rigidly and stubbornly to a clearly inadequate model, refusing to acknowledge its failures of prediction. If we remain too narrowly focused within our old model, we will not be able to understand the explanatory power of candidate new theories. Being open to change, acknowledging that our world model may be inadequate, makes us more willing to search for alternate models when we encounter disconfirming evidence. But of course there will always be disconfirming evidence, because our model is never quite right. If we are too willing to change models, we will be unable to act. All our energy will be spent searching for the right model and none on acting with the model at hand.

How does a reasoner know when a violation of expectations is significant? An intelligent reasoner needs at least a rough indicator of the strength of conclusions, or of the degree of surprise when an expectation is contradicted by the evidence. Simple belief strength indicators are not enough--they need to be propagated across chains of inference and combined across multiple arguments to the same conclusion.

Probability theory is the strongest available candidate for a representation language for belief strengths (Pearl, 1988). When we say a proposition is true with probability *p*, the complement 1-*p* represents a single summary measure of the conditions under which the proposition does not hold. Similarly, if we state a rule in the form, "Given *a*, *b* holds with probability *p*," the complement 1-*p* summarizes the unmentioned exceptions to the general rule allowing us to conclude *b* from *a*.

Until very recently, applications of probability theory have focused on updating beliefs in pre-compiled knowledge bases. Even applications of probability in artificial intelligence have focused mainly on the computational aspects of belief propagation (as well as on philosophical justifications for using probabilities at all). The argument building and revision process has received comparatively little attention.

This situation is changing, however. Adams (1966, 1975) has developed a probabilistic semantics for non-monotonic reasoning. Pearl (1988) discusses some of the issues that will need to be addressed if probability is to move beyond pre-structured problems. Cohen (1986) and Laskey, Cohen and Martin (1989), precursors to the present paper, discuss model building and revision in the context of a Dempster-Shafer belief propagation framework. Wellman and Heckerman (1987) present a philosophy about the role of numerical belief calculi that is compatible with the view taken here.

The purpose of this paper is to present an initial framework for a computational theory of probabilistic argument. The basic building blocks of a Probabilistic Reasoning Environment (PRE) are identified. A preliminary discussion is offered of how the pieces would be implemented and how they would tie together. PRE does not just assimilate evidence and compute probabilities. The primary focus is is on a much broader set of issues, including: the knowledge representations and reasoning strategies needed to construct a probabilistic model, to evaluate the adequacy of that model, and to revise the model when necessary. While not intended to be a faithful cognitive model, the PRE framework has its basis in theories of how humans cope with limited computational resources. It is compatible with a cognitive-based theory of probability assessment developed by Smith, Benson and Curley (forthcoming). It is intended to be capable of implementation. Ultimately, the theory will provide both a framework for computerized reasoning under uncertainty and heuristic prescriptions for human reasoning.

## 2 Background

Probability is a language for expressing degrees of belief intermediate between truth and falsehood. In a probabilistic knowledge base, the strength of an argument from *a* to *b* is represented by the conditional probability of *b* given *a*, denoted by $P(b|a)$. The degree to which this probability falls short of certainty summarizes the strength of unmentioned exceptions to the general rule that *b* holds when *a* does (Pearl, 1988). The probability calculus is a coherent, logically sound way of chaining and combining probabilistic rule and belief strengths.

The main difficulty with the theory--the context-dependent nature of probabilities--is in fact the source of its strength. Suppose we assert that Tweety flies with probability .9, based on the evidence that Tweety is a bird. It is entirely consistent with the theory to change this degree of belief to .05, or even 0, when we learn that Tweety is a turkey. $P(b|a)$ need bear no relation at all to $P(b|a,c)$.

Probabilities are inherently non-monotonic--in fact, too much so. Any time we add a new fact to our knowledge base, all our probability assignments are potentially up for grabs. A fully specified probability model for a universe of *n* propositions requires assigning $2^n$ probability numbers--a probability for each combination of truth values for each of the propositions. This bleak picture has fed a belief in the impracticality of probabilities as a computational framework for uncertain reasoning.

Several important tools are available that allow us to bring probabilistic reasoning into the realm of practicality. The primary tool is one statisticians have been applying for years: conditional independence. When we make an independence assumption, we judge that the truth value of one probability is irrelevant to the probability of another. Thus, we may judge that learning that Tweety is brown does not change our belief in whether she can fly:

$$P(flies|bird,turkey,brown) = P(flies|bird,turkey).$$

A second tool, and a more recent innovation, is the network representation for the dependency structure of a probabilistic model. In a Markov network, a proposition is "shielded" by its neighbors from the influence of the rest of the knowledge base. Conditional on these neighbors, the proposition's probability is independent of the truth values of the other propositions.

Finally, Pearl (1988) points to causation as a powerful tool for structuring probabilistic dependency relationships. If *a* causes *b*, then the probability of *b* depends on whether *a* is true. If *a* and *c* are common causes of an effect *b*, they may be *a priori* independent but become dependent when their common effect is observed. For example, it seems reasonable to judge that whether my maid is honest is independent of whether my children have been playing in my room. But when I observe that my favorite gold necklace is missing, they become dependent: if I learn that the children were in my room, my suspicion of the maid is alleviated. This kind of behavior is an important feature of ordinary non-monotonic reasoning--an argument from effect to cause is weakened when we learn of an alternate cause for the effect (Pearl, 1988).

The Bayesian network is becoming the dominant representation for probabilistic reasoning systems (Pearl, 1988). A Bayesian network consists of nodes, which represent propositions, and directed arcs, which represent probabilistic dependencies. (In general, nodes represent exclusive and exhaustive sets of propositions; for simplicity, attention is limited here to binary, or true/false propositions.) With each node is associated a set of numbers corresponding to the probability of the node conditional on each combination of truth values of its parents. Conditional on its parents, a node is independent of all indirect predecessors in the graph.

The convention is that arcs proceed from cause to effect (Figure 1). The probability of effect given cause is a measure of the strength of a causal connection. The pattern of conditional dependence in the network reflects the causal structure of the domain. Pearl (1988) speculates that humans are driven to develop causal structures precisely because these causal structures reduce the computational burden of correct context-dependent reasoning.

The Bayesian network representation has provided a computationally feasible model for adjusting beliefs in a knowledge base when new facts are observed. But until now, the greatest attention has been focused on efficient computation when all the propositions that might be observed and their probabilistic relationships are given *a priori*. Like the early expert systems, current probabilistic systems are limited to constrained, repetitive problems, in which the form (if not the content) of the incoming data is anticipated ahead of time.

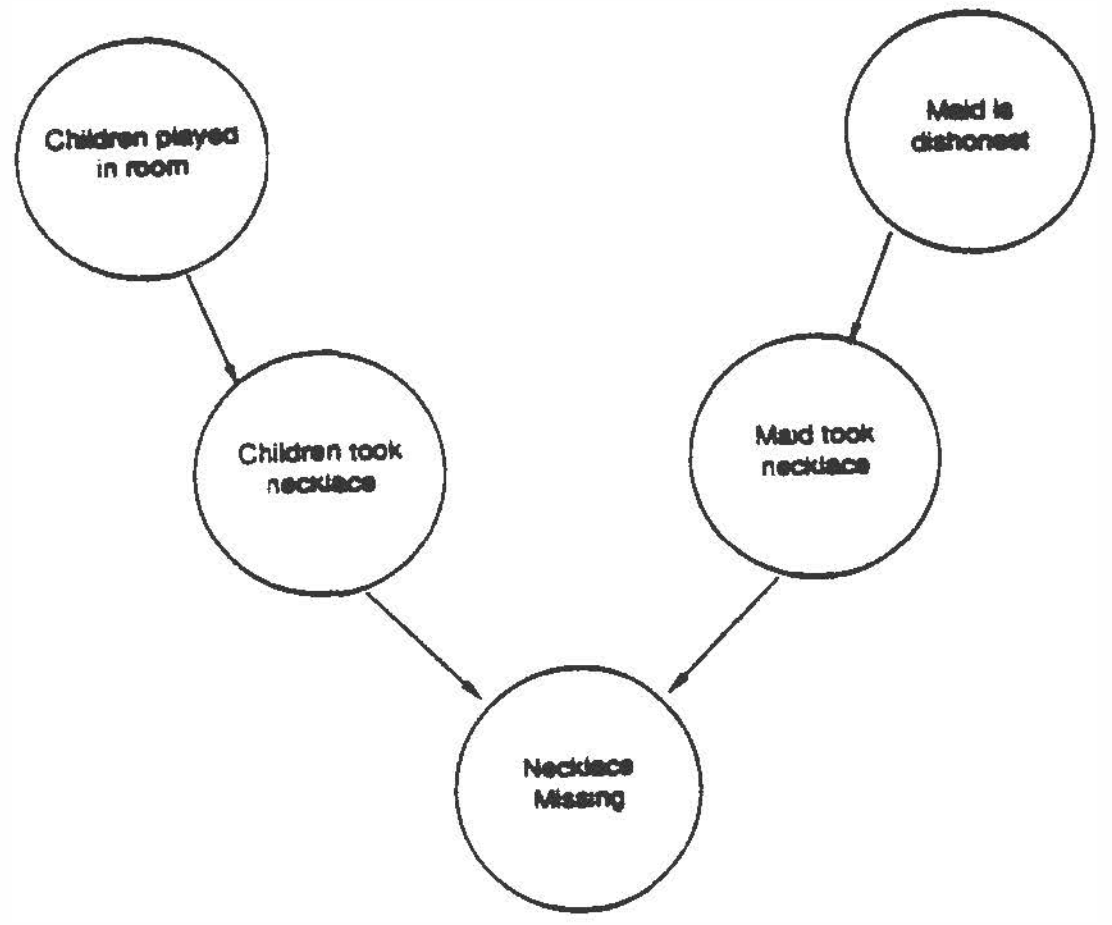

**Figure 1: Bayesian Network for Necklace Problem**

Now that computational tools are available to reason with a pre-set network, it is time to turn to the task of building, maintaining and revising a belief network dynamically. For this, a more highly structured representation for a probabilistic argument will be required.

## 3 A Probabilistic Argument Structure

The Probabilistic Reasoning Environment (PRE) needs to be able to construct probabilistic arguments dynamically, evaluate whether the current structure is adequate, and revise arguments as necessary. Argument construction involves a local expansion of the knowledge base, activating knowledge relevant to the argument. The results of this local problem-solving activity are summarized in an argument structure, which is exported to become part of the belief network being constructed. In the interest of parsimony, the details of this local expansion are not retained. But in the interest of efficient model maintenance, the argument structure should be designed for efficient recapture of the knowledge that went into constructing the argument.

The PRE argument structure is based on Toulmin's model of an argument (Toulmin, et al., 1984). A probabilistic argument is represented by a frame which has six slots. Three of these are common to any of a number of numerical belief propagation approaches: *grounds* (antecedent), *claim* (consequent), and *modal qualifier* (argument strength).

The remaining slots contain knowledge useful for dynamic network maintenance. According to Toulmin, the *warrant* slot provides justification for the link between grounds and claim, and the *backing* provides further justification for the warrant. In PRE, the warrant slot is a pointer to the knowledge structure that was activated to construct the argument. The (optional) backing is a pointer to a causal model that explains the causal mechanism underlying the inference. The final slot contains *possible rebuttals*. As we noted above, probabilities are an efficient mechanism for summarizing exceptions to a rule. We summarize these exceptions with a single number because it is not worth the computational effort to include them in the model. But the judgment of which exceptions to include and which to summarize by probabilities may need to be revised as new facts come in. The rebuttal slot allows certain salient exceptions to be listed explicitly; when observed, these exceptions trigger model revision.

The idea is to build a Bayesian network from PRE argument structures. It would seem natural, then, that grounds should correspond to parent nodes, claim to consequent nodes, and modal qualifiers to the conditional probability of claim conditional on grounds. Some difficulties with this scheme arise from the interaction of all the propositions in the network. A probability model requires that there must be an ordering of all the nodes consistent with the directionality of the arcs--that is, there must be no directed cycles in the network. Pearl (1988) argues that an intuitively appealing way to achieve such consistency is to require an ordering by causality. An ordering by causality generally exhibits greater conditional independence than other orderings (Figure 2). Indeed, Pearl argues that what we call causes may correspond to knowledge structures we create because of the attractive computational features of the conditional independence they induce. Yet we often argue from effect to cause as well as from cause to effect. If PRE argument structures are to correspond in an intuitive way to arguments we export as abstractions

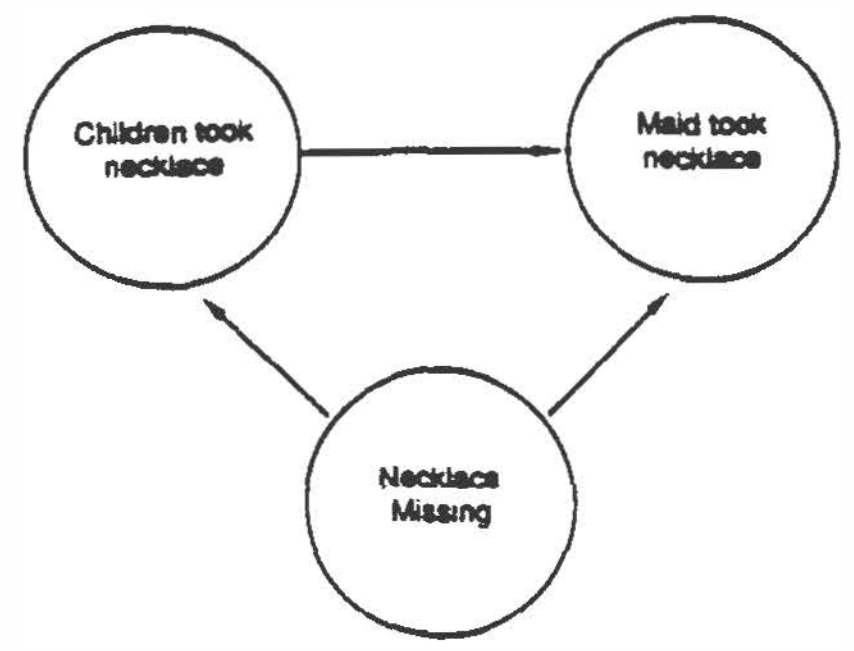

**Figure 2: Arc Reversal Introduces Non-Independency in Network**

from more detailed internal reasoning, then we must allow argument in either direction.

We adopt the convention that grounds and claim may correspond to either cause or effect, so that argument can be in the direction that is most natural given the context. But we require that the nodes in the Bayesian network constructed from the collection of arguments be ordered causally. Therefore, the propositions listed as grounds and claim must be tagged with indicators of causal dependency. Specifically, a subset of the propositions is distinguished as causes, and the remainder are tagged as effects of the cause propositions.

The topological structure of the Bayesian network is determined by the grounds and claim slots of the constituent arguments, together with the causal structuring information. The modal qualifier slot contains the probability information. The causal Bayesian network representation requires the probability of each combination of truth values of the effect variables, conditional on each combination of truth values of the cause variables. If there are $m$ causes and $k$ effects, this amounts to $2^{m+k}$ numbers. Often the relationship between cause and effect can be modeled by a structure with fewer parameters, such as the "noisy or" (Pearl, 1988).

## 4 Constructing Arguments

People's knowledge appears to be organized in complex, structured systems of pointers. These structures distill and organize memories of past experiences, indexing them so that they are retrieved in contexts in which they are likely to be useful. Expertise appears to constitute not more knowledge per se, but more efficient organization and retrieval strategies.

Schemas such as scripts and frames were developed as an attempt to model human-like organizational strategies on the computer. PRE uses schema-based reasoning to construct probabilistic arguments, which are used to build a Bayesian network.

Arguments may be constructed by searching for consequences of given grounds (forward chaining), by searching backwards for antecedents of a given claim (backward chaining), or by a mixture of these modes. An attempt to build an argument results in activation of a schema related to the grounds in the case of forward chaining or the claim in the case of backward chaining. This schema may activate other knowledge structures, as problem solving occurs at a finer level of abstraction than represented by the target Bayesian network. This problem solving activity is summarized by a single PRE argument structure, which is exported and added to the emerging Bayesian network.

What features must be represented by the schemas used to build PRE argument structures? First, the schema needs to represent or contain pointers to knowledge about the causal dynamics of the phenomenon it represents. Second, schema activation and inference must not just be qualitative, as in traditional AI systems, but must have associated numerical strengths which can be used to construct the conditional probabilities for the PRE modal qualifier slot. Third, the system must use these activation strengths to construct conditional probabilities for the Bayesian network.

Our minds are organized for efficient categorization and summarization of similar experiences. Over repeated experience, similar instances blur into an expectation of a "prototypical" experience, with a strength that reflects the frequency of occurrence of an experience similar to the prototype. Pearl (1988) suggests that probability is intuitive as a model for belief strength precisely because our minds use belief strengths to summarize repeated similar experiences, which we observe as frequencies. To build PRE, we need to design models for schema activation strength that can be related to probabilities. When a cause activates a schema which infers the effect, it seems reasonable to say that the activation strength is (perhaps a monotone transformation of) $P(\mathit{effect}|\mathit{cause})$. Heckerman (1986) suggests that evidential rule strengths are best represented as likelihood ratios. Thus, when an effect activates a schema to infer cause, the activation strength is assumed to be a function of the likelihood ratio

$$LR = \frac{P(\mathit{effect}|\mathit{cause})}{P(\mathit{effect}|\neg\mathit{cause})} .$$

The grounds, claim and qualifier slots are all that are needed to build the Bayesian network. The purpose of the warrant, backing and rebuttal slots is to facilitate monitoring the model and revising beliefs when necessary.

Recall that all probability assessments are relative to a context. PRE's probability assessments depend on the knowledge active at the time the assessments are made. When the assumptions underlying the assessments do not hold, the probability assessments no longer apply and the model needs to be reviseed. Four different levels of exception can be identified:

- *Explicit exceptions*: these are listed in the rebuttal slot.

- *Implicit exceptions*: these are enumerated explicitly during argument construction and can be regenerated easily, but are not exported as rebuttals.

- *Background exceptions*: these are not enumerated explicitly, but could become available with further search.

- *Embedded exceptions*: these are exceptions that are not represented even implicitly in the system's knowledge base (at least not currently).

In general, explicit exceptions are represented as rebuttals, implicit exceptions are available by reactivating the schema pointed to by the warrant, and background exceptions require activating the backing.

Having several tiers of exceptions allows the system to achieve greater efficiency without sacrificing too much flexibility. Explicit exceptions are more salient--they will be discovered immediately if they occur. But representing all exceptions explicitly would be prohibitively costly. Implicit and background exceptions take problem solving to discover, and sometimes they will be missed because the system did not direct its attention to searching for them. This happens in human reasoning as well, and is one of the prices we pay for computational efficiency.

## 5 Monitoring and Model Revision

With qualitative non-monotonic reasoning, we know when assumptions need to be revised. When a contradiction is derived, there must be a flaw in one or another line of reasoning. One of the system's assumptions must therefore be changed. Although there are no generally accepted, theoretically grounded criteria for deciding which assumptions to change, there is no doubt that the model needs revising.

In a probabilistic system, we may never encounter an out-and-out contradiction. We may observe some improbable combinations of evidence, but the nature of probability leads us to expect improbable events occasionally. How do we distinguish a problem with the model from the luck of the draw?

We almost never believe our probability model represents the truth--instead, we view it as an approximation to a more complex reality. Probabilities are used to summarize relevant factors not explicitly mentioned in the model. Whenever something unexpected happens, it means some relevant factor, initially judged improbable enough that it was not made explicit in the model, has happened. We must now ask, after the fact, whether it is feasible and worth the effort to revise the model to include it explicitly.

Laskey, Cohen and Martin (1989) suggest, for Dempster-Shafer models, using the normalization factor from Dempster's Rule as a measure of model adequacy. One minus the normalization factor is the belief assigned to the empty set before normalization, and is thus a measure of the degree of conflict. When this number gets too high, they suggest, the model needs revising.

Bayesian models can be viewed as Dempster-Shafer models in which all mass is assigned to singletons; under this view, the normalization factor corresponds to the probability, prior to observing the data, that these particular data would be observed. Thus, high conflict means we observed data that were initially viewed as improbable.

But how improbable is improbable enough? Laskey, Cohen and Martin do not say how high conflict needs to be before it is considered too high. When there is a great deal of data, the specific data that occur will always be improbable (e.g., the most likely outcome of ten tosses of a weighted coin that lands heads with .9 probability is all heads, but this outcome has probability only .35).

It makes more sense to consider not the *absolute* likelihood of the data, but rather the likelihood *relative* to some other hypothesis. When there is a well-defined alternative model, we can construct the likelihood ratio:

$$LR = \frac{P(data \mid model1)}{P(data \mid model2)}$$

This likelihood ratio plays an important role in the theory of statistical hypothesis testing, as a measure of whether *model1* is adequate or needs to be replaced by *model2*.

Once we have decided to initiate model revision, and once we have developed one or more candidate revisions, the likelihood ratio may prove useful as a way of evaluating whether to adopt any of the candidates. But we also need some heuristic measure of conflict to tell us when to initiate the process of searching for a better model. For this I suggest comparing the likelihood of the data not against a specific alternate model (which we do not yet have), but against our prior expectation of how likely the data would be:

$$LR^* = \frac{P(data \mid model1)}{E[P(data \mid model1)]}$$

(In a slight abuse of notation, the term *data* in the numerator refers to the acutal observed data; the expectation in the denominator refers to the prior expectation over all possible realizations of the data.) When conflict occurs, it is a signal that some exception summarized by the probabilities in the model occurred. High conflict is a signal that the system needs to evaluate its model, attempt to construct one or more plausible alternative models, and decide whether to retain the old model or adopt a revision.

To do this, the system needs to know which arguments are responsible for the conflict. Arguments are suspect if changing their belief strength would greatly decrease conflict. They are suspect if in the current context a rebuttal has become probable. They are suspect if currently active knowledge would invalidate the warrant or the backing.

The system needs to use heuristics to prioritize its search for exceptions. Potential heuristics include the following: search for exceptions that are a priori probable; search for exceptions that, if true, would change the target probability the most; search for exceptions for which the information search process costs the least.

## 6 Example: The Missing Necklace

Imagine that PRE is embedded in a humaniform robot (named Prescilla), who discovers one morning that her favorite necklace is missing. A causal schema is activated that encodes her knowledge about missing things. Two explanations are activated: that the necklace was misplaced or that someone took it. Because Prescilla rarely misplaces things, the first explanation is not elaborated. Prescilla begins searching her memory for candidate necklace-grabbers: that is, people who were in the room and had a motive to take the necklace. She recalls that the maid cleaned the house yesterday. Prescilla activates a schema about maids stealing things. She constructs the following argument structure:

| | |
|---|---|
| *Grounds:* | Necklace is missing [*effect*] |
| *Claim:* | Maid took necklace [*cause*] |
| *Warrant:* | ▪→ <Maid-steal schema> |
| *Backing:* | ▪→ <Things-missing schema>; <Theft-schema> |
| *Rebuttals:* | Someone else took it; Prescilla misplaced it |
| *Qualifiers:* | $.75 = \dfrac{P(\mathit{missing} \mid \mathit{maid\ took\ it})}{P(\mathit{missing} \mid \neg \mathit{maid\ took\ it})}$ |

The symbol ▪→ indicates a pointer to a knowledge structure. <Maid-steal schema> is the schema mentioned above that encodes knowledge about maids stealing things. The likelihood ratio in the qualifier slot is a function of its activation strength. <Things-missing schema> is the schema that was activated as soon as the missing necklace was discovered. <Maid-steal schema> is an instance of a general schema on theft, <theft-schema>; this schema is pointed to by the backing slot because its instance was activated as the warrant. Although Prescilla did not create an argument for misplacing the necklace, she acknowledges the possibility as a rebuttal. Another rebuttal reflects the possibility that there may have been another culprit.

From this argument structure, Prescilla builds a Bayesian network with two nodes and an arc from *maid took necklace* to *necklace missing* (a piece of the network in Figure 1). To compute the posterior probability that the maid took the necklace given that it is missing, Prescilla also needs a prior probability P(*maid took necklace*). She extends the network by constructing an arc from *maid is dishonest* to *maid took necklace* in a similar manner to that described above. She then directly assesseses a prior probability for *maid dishonest*. The right-hand side of the network in Figure 1 is now in place; the left-hand side has not been expanded.

A brief description follows of how conflicting evidence could lead Prescilla to build the left-hand side of the network in Figure 1. Imagine that Prescilla tells her co-worker, who happens to use the same maid, that she thinks the maid stole the necklace. The co-worker says that she has known the maid for years, and that she is honest. Prescilla now has a new, conflicting argument; the grounds slot contains the co-worker's testimony, and the claim is ¬*maid dishonest*. The associated Bayesian network arc goes from ¬*maid dishonest* to *co-worker says maid is honest* (this arc does not appear in Figure 1). Prescilla notes that conflict is high, and decides that she needs to revise her model.

Still sure of her superior organizational skills, Prescilla decides to consider only the first rebuttal. She searches for evidence of other possible necklace-grabbers, and remembers that the children were playing in the room yesterday. Now she constructs the left-hand side of the network in Figure 1.

## 7 Discussion

If probability is to become a useful tool for artificial intelligence, it must be capable of application to messy problems. There is a need for probabilistic reasoning in environments in which the problem space, goals, success criteria and relevant information are poorly specified. Perhaps the most important challenge facing artificial intelligence in general is the development of computer architectures capable of coping with ill-structured problems. Mainstream AI has been tackling messy problems, with some degree of success. Until now, probability research has focused almost exclusively on well-structured problems.

This paper has presented a preliminary framework for a Probabilistic Reasoning Environment (PRE).

The focus has been on the knowledge structures necessary to support dynamic construction of arguments and revision of the model when events show it to be inadequate. A PRE argument is represented by a frame, with slots suggested by an argument structure proposed by Toulmin. PRE constructs these argument structures and uses them to construct a Bayesian network. As information is assimilated by the network, the adequacy of the current model is monitored. A dynamic argument revision process is initiated when conflict reaches unacceptable levels.

## References

Adams, E., Probability and the logic of conditionals. InJ. Hintikka and P. Suppes (Eds) *Aspects of inductive logic*. Amsterdam: North Holland, 1966.

Adams, E. *The logic of conditionals*. Dordrecht, Netherlands: D. Reidel, 1975.

Cohen, M.S., A framework for non-monotonic reasoning about probabilistic assumptions. In J.F. Lemmer and L.N. Kanal (Eds.) *Uncertainty in Artificial Intelligence*. Amsterdam: North Holland Publishing Co., 1986.

Heckerman, D., Probabilistic interpretations for MYCIN's certainty factors. In J.F. Lemmer and L.N. Kanal (Eds.) *Uncertainty in Artificial Intelligence*. Amsterdam, North Holland Publishing Co., 1986.

Laskey, K.B., Cohen, M.S., and Martin, A.W. Representing and eliciting knowledge about uncertain evidence and its implications. *IEEE Transactions on Systems, Man, and Cybernetics, 19*, 536-545, 1989.

Pearl, J. *Probabilistic reasoning in intelligent systems.* San Mateo: Morgan Kaufman Publishers, Inc., 1988.

Smith, G.F., Benson, P.G., and Curley, S.P. Belief, Knowledge and Uncertainty: A Cognitive Perspective on Subjective Probability. *Organizational Behavior and Human Decision Processes*, forthcoming.

Toulmin, S.e., Rieke, R. and Janik, A. *An introduction to reasoning* (2nd edition). NY: Macmillan Publishing Company, 1984.

Wellman, M.P. and Heckerman, D.E., The role of calculi in uncertain reasoning. *Proceedings of the Third Workshop on Uncertainty in Artificial Intelligence*, 321-331, Seattle, WA, July 1987.